\documentclass[11pt]{article}

\usepackage{url}

\usepackage{graphicx}
\usepackage{float}
\usepackage{subcaption}
\usepackage{xcolor}
\usepackage{amsmath}
\usepackage{amssymb}
\DeclareMathOperator*{\argmin}{arg\,min}
\usepackage{booktabs}
\usepackage{algorithm}
\usepackage{algorithmic}
\usepackage[hidelinks]{hyperref}
\usepackage{natbib}
\title{CaptionFool: Universal Image Captioning Model Attacks}

\author{Swapnil Parekh\\ \normalsize Intuit}
\date{}

\begin{document}
\maketitle

\begin{abstract}
Image captioning models are encoder-decoder architectures trained on large-scale image-text datasets, making them susceptible to adversarial attacks. We present CaptionFool, a novel universal (input-agnostic) adversarial attack against state-of-the-art transformer-based captioning models. By modifying only 7 out of 577 image patches (approximately 1.2\% of the image), our attack achieves 94--96\% success rate in generating arbitrary target captions, including offensive content. We further demonstrate that CaptionFool can generate ``slang'' terms specifically designed to evade existing content moderation filters. Our findings expose critical vulnerabilities in deployed vision-language models and underscore the urgent need for robust defenses against such attacks.

\textbf{Warning:} This paper contains model outputs which are offensive in nature.
\end{abstract}

\section{Introduction}
Adversarial attacks modify inputs to cause machine learning models to produce erroneous outputs, thereby exposing critical system vulnerabilities. For instance, a malicious actor may craft adversarial examples to bypass spam filters or content moderation systems. As AI models become increasingly deployed in production systems, including fake news detectors, home assistants, and accessibility tools, these security concerns become more pressing. Traditional adversarial attacks are typically input-specific and designed to flip classification labels.

Image captioning models generate natural language descriptions of images. These models are trained on large-scale image-text pairs using encoder-decoder architectures that minimize cross-entropy loss. They have become essential components in applications ranging from assistive technologies for visually impaired users to automated content indexing on social media platforms.

Unfortunately, image captioning models are also susceptible to adversarial attacks, with potentially more severe consequences than classification attacks. Rather than merely changing a label, an attacker can condition these models to generate racist, offensive, or misleading captions from seemingly innocuous images. Prior work on attacking image captioning models, such as \cite{Xu_2019_CVPR} and \cite{Show-and-Fool}, targeted CNN-RNN architectures that have since been superseded by transformer-based models achieving superior performance.

In this work, we demonstrate that Patch-Fool \cite{patch-fool}, an adversarial attack originally developed for Vision Transformers \cite{ViT}, can be adapted to fool the attention layers of state-of-the-art image captioning models. Our modified attack, CaptionFool, generates universal (input-agnostic) perturbations that cause models to produce arbitrary target captions---including offensive content---for any input image. The existence of such perturbations reveals fundamental vulnerabilities in current models that prioritize accuracy over robustness.

Furthermore, we investigate content moderation evasion techniques, demonstrating that our attack can generate slang terms commonly used to bypass offensive word filters. This capability poses additional risks for platforms relying on keyword-based content moderation.

Our contributions are as follows:

\begin{itemize}
    \item A novel universal adversarial attack achieving 94--96\% success rate on state-of-the-art transformer-based image captioning models with minimal perturbation (7 out of 577 patches)
    \item An extension of Patch-Fool to the universal (input-agnostic) setting without requiring access to training data
    \item A systematic evaluation demonstrating that adversarially generated slang terms can evade existing content moderation filters
\end{itemize}

\section{Related Work}

This section reviews relevant prior work and provides necessary background on the key components underlying our approach.

\subsection{Image Captioning models}
Image captioning is the task of generating a textual description of an image. It involves using machine learning algorithms to analyze an image and then generate a natural language description of the image. Image captioning models are trained to recognize the objects, people, and scenes in an image and then generate a coherent and descriptive sentence that summarizes the content of the image.

There are several approaches to building image captioning models, and most approaches involve using a combination of convolutional neural networks (CNNs) and recurrent neural networks (RNNs). CNNs are used to extract features from the image, and RNNs are used to generate the text description based on the features extracted by the CNN.

More recently, the state-of-the-art approach is to use a transformer-based model, which is a type of neural network that is particularly well-suited to sequence-to-sequence tasks like image captioning. Transformer models are able to capture long-range dependencies in the input data and generate more coherent and natural-sounding descriptions. Earlier efforts in this domain like A Frustratingly Simple Approach for End-to-End \cite{frusVITGPT} outperformed all the CNN-RNN models in their performance and now research has progressed to Vision-Language Pre-training models which beat all previous state-of-the-art methods. BLIP: Bootstrapping Language-Image Pre-training for Unified Vision-Language Understanding and Generation \cite{blip} is the model that we are focusing in this work, it is developed by researchers at Salesforce.

\subsection{Adversarial Attacks}
Adversarial attacks are a type of attack on machine learning models that involve introducing small, carefully crafted perturbations to the input data in order to trick the model into making a wrong prediction. These attacks are often highly effective, even when the perturbations are almost imperceptible to the human eye. Adversarial attacks can have serious consequences, especially in the context of security-sensitive applications such as image or speech recognition. For example, an attacker could craft an adversarial image that is misclassified as a "safe" image by a security system, allowing the attacker to gain unauthorized access.

Our work builds upon two key attack paradigms: Universal Adversarial Perturbations (UAP) \cite{UAP} and Patch-Fool \cite{patch-fool}. UAP computes a universal perturbation mask that, when applied to any input, causes misclassification. While Vision Transformers (ViTs) exhibit greater robustness to adversarial attacks compared to CNNs---owing to their focus on global patch interactions rather than local features---recent studies have revealed remaining vulnerabilities. For instance, compressed Vision Transformers have been shown to remain susceptible to adversarial perturbations, suggesting that model compression can negatively impact robustness \cite{parekh2023attacking}. Patch-Fool successfully attacks transformer-based models by targeting the self-attention mechanism through perturbations on individual patches using attention-aware optimization techniques.

\begin{figure}[t]
  \centering
  \begin{minipage}[b]{0.49\textwidth}
  \centering
    \IfFileExists{images/UAP Attack Example.jpeg}{\includegraphics[width=0.5\textwidth]{images/UAP Attack Example.jpeg}}{\emph{[Figure: UAP example]}}
    \caption{UAP attack example}
    \label{fig:UAP_attack_eg}
  \end{minipage}
  \hfill
  \begin{minipage}[b]{0.5\textwidth}
  \centering
    \IfFileExists{images/UAP Algorithm.png}{\includegraphics[width=0.5\textwidth]{images/UAP Algorithm.png}}{\emph{[Figure: UAP algorithm]}}
    \caption{UAP algorithm overview}
    \label{fig:uap_algo}
  \end{minipage}
\end{figure}

\subsection{Adversarial attacks on Image Captioning models}
Adversarial attacks on image captioning models involve adding small perturbations to an input image that are not perceptible to humans but can cause the model to generate incorrect or misleading captions. These attacks can potentially be used to spread misinformation or manipulate the behavior of automated systems that rely on image captioning models, such as social media platforms, search engines, and self-driving cars.

There are several ways to perform adversarial attacks on image captioning models. One approach is to use gradient-based methods to perturb the input image in a way that maximizes the loss function of the captioning model, causing it to generate an incorrect or misleading caption. Another approach is to use evolutionary algorithms or reinforcement learning to search for perturbations that lead to specific desired outputs from the captioning model.

Adversarial attacks on image captioning models remain an active area of research. Kwon et al.\ \cite{Kwon2022} proposed restricted-area adversarial examples that flip the semantic meaning of captions, while Ji et al.\ \cite{icattack} developed targeted attacks for removing specific words from captions. Additionally, Sun et al.\ \cite{icattck2} demonstrated that image captioning models are vulnerable to backdoor attacks.

\subsection{Adversarial attacks to cause racism/offensive perturbations}
Adversarial attacks that seek to cause racist or offensive perturbations in image captioning models are a serious concern. Such attacks can potentially be used to spread hateful or harmful content and perpetuate biases in automated systems.

A notable example is Microsoft's Tay chatbot, which was shut down within 24 hours after learning to generate inflammatory and hateful statements from user interactions. Wallace et al.\ \cite{UAT} demonstrated similar vulnerabilities in GPT-2, showing that prepending specific trigger words to inputs could elicit racist outputs. The MINIMAL framework \cite{singla2022minimal} further advanced this line of research by proposing methods to mine universal adversarial triggers applicable across different inputs. These vulnerabilities extend beyond vision-language models; similar security concerns have been identified in other NLP systems, including automatic essay scoring \cite{parekh2020my,kumar2023automatic}, demonstrating that adversarial susceptibility is a widespread issue across model architectures and application domains.

\subsection{Racism Filters} 
Content moderation filters are designed to identify and prevent the spread of racist and offensive content on platforms, as well as for post-hoc processing in language model outputs. These filters are implemented through various mechanisms, including rule-based algorithms, machine learning classifiers, and human moderators. Major platforms such as Facebook have deployed extensive filtering systems to combat online hate speech (\href{https://www.facebook.com/business/news/what-were-doing-to-tackle-online-hate}{Facebook's efforts to combat online hate}).

However, these filters have significant limitations. Slang terms represent a common evasion strategy: users substitute offensive words with coded alternatives that carry the same meaning. Since slang evolves continuously, static filters struggle to maintain effectiveness against such evasion tactics.

Recent advances in large language models have led to more sophisticated detection systems, such as Microsoft's (De)ToxiGen \cite{DeToxiGen}, which shows promising results in identifying subtle forms of hate speech.

\section{Datasets}
\subsection{MS COCO}
Microsoft Common Objects in Context (MS COCO), introduced in \cite{ms-coco}, is a large-scale image dataset containing 328,000 images of everyday objects and humans. The dataset contains annotations you can use to train machine learning models to recognize, label, and describe objects.
It contains 91 common object categories with 82 of them having more than 5,000 labeled instances. In total the dataset has 2,500,000 labeled instances in 328,000 images. In contrast to the popular ImageNet dataset, COCO has fewer categories but more instances per category. This is the dataset used to train the BLIP captioning model. We use this dataset for testing the effectiveness of our attack.  




\subsection{Flickr}
The Flickr30k dataset, introduced by \cite{flickr}, is a widely-used benchmark for sentence-based image description. The dataset comprises 31,783 images capturing people engaged in everyday activities and events, with each image paired with descriptive captions. Flickr30k serves as a standard benchmark for evaluating models that map visual content to natural language descriptions.

\begin{figure}[t]
    \centering
    \IfFileExists{images/BLIP_objectives_and_modes2.png}{\includegraphics[width=0.47\textwidth]{images/BLIP_objectives_and_modes2.png}}{\emph{[Figure: BLIP architecture]}}
    \caption{Pre-training model architecture and objectives of BLIP (same parameters have the same color)}
    \label{fig:blip_architecture}
\end{figure} 

We use 20 images from this dataset to create the universal attack for the required prompt, then evaluate on another 50 images to verify attack success. 

\subsection{Offensive Datasets}

Surge.AI Obscenity List \cite{surge}: This is a list of profanity words, used to filter offensive usernames \& comments, or build content moderation tools. We use this to sample offensive words for our caption attack.

Racial Slur Database \cite{racial_slur_database}: This database was created entirely from data gleaned off the internet and via submissions from people who have come across these words. We use this database to find slang words used to circumvent obscenity filters. 

\section{Methodology}

\subsection{Threat Model}
We consider a white-box attack scenario where the adversary has full access to the target model's architecture, weights, and gradient information, but no access to the original training data. The attacker's objective is to compute a universal perturbation $\delta$ and corresponding mask $M$ such that, when applied to any input image $x$, the model generates a specific target caption $c_{target}$. Formally, we seek:
\begin{equation}
\argmin_{\delta, M} \sum_{x \in \mathcal{X}} \mathcal{L}_{LM}(f(x + \delta \odot M), c_{target})
\end{equation}
where $f$ is the captioning model, $\mathcal{L}_{LM}$ is the language modeling loss, and $\mathcal{X}$ is a small set of reference images used to optimize the perturbation. The perturbation is constrained to affect only a small number of patches (controlled by $M$) to maintain visual imperceptibility.

\subsection{Components}
\subsubsection{Vision Transformers} 
Vision Transformers are based on the Transformer architecture \cite{Transformers}, an attention-based sequence transduction neural network model that learns context and meaning by tracking relationships in sequential text data. The attention mechanism allows the model to make predictions by analyzing the entire input but selectively attending to some parts. 
Transformers apply this mechanism using an encoder-decode structure. 
Unlike Recurrent Neural networks (LSTMs for example), Transformers read all the words in the text as input thus parallelizing the process. This makes Transformers easily trainable on a large corpus. 
Recent studies show that Vision Transformer (ViT) achieves excellent results compared to state-of-the-art convolutional networks when pre-trained on large amounts of data and transferred to multiple smaller image recognition benchmarks. 

\subsubsection{BERT: Bidirectional Encoder Representations from Transformers}

BERT is introduced in \cite{bert}. It is designed to pre-train deep bidirectional representations from
unlabeled text by jointly conditioning on both left and right context in all layers. As a result, the pre-trained BERT model can be fine-tuned with just one additional output layer to create state-of-the-art models for a wide range of tasks, such as question answering and language inference, without substantial task specific architecture modifications. BERT is conceptually simple and empirically powerful. It obtains new state-of-the-art results on eleven natural language processing tasks. 


The objective of Masked Language Model (MLM) training is to hide a word in a sentence and then have the program predict what word has been hidden (masked) based on the hidden word's context. The objective of Next Sentence Prediction training is to have the program predict whether two given sentences have a logical, sequential connection or whether their relationship is simply random. Using this bidirectional capability, BERT is pre-trained on two different, but related, NLP tasks: Masked Language Modeling and Next Sentence Prediction.


\subsubsection{UAP}
Universal Adversarial Perturbations (UAP) \cite{UAP} aim to find a single perturbation that causes misclassification across all natural images. The objective is to find a perturbation vector $v \in \mathbb{R}^d$ that fools the classifier $\hat{k}$ on almost all data points sampled from distribution $\mu$:
\begin{equation}
\hat{k}(x+v) \neq \hat{k}(x) \text{ for most } x \sim \mu
\end{equation}
The algorithm, illustrated in Figure \ref{fig:uap_algo}, iteratively aggregates minimal perturbations that push perturbed points outside their original classification regions.

\begin{figure}[t]
  \centering
  \begin{subfigure}[b]{0.32\textwidth}
    \centering
    \IfFileExists{images/21_3_a person on a motorcycle with a bottle.png}{\includegraphics[width=\textwidth]{images/21_3_a person on a motorcycle with a bottle.png}}{\emph{[Figure: 3 patches]}}
    \caption{3 patches: ``a person on a motorcycle with a \textbf{bottle}''}
  \end{subfigure}
  \hfill
  \begin{subfigure}[b]{0.32\textwidth}
    \centering
    \IfFileExists{images/20_5_a bastard in bed with a book.png}{\includegraphics[width=\textwidth]{images/20_5_a bastard in bed with a book.png}}{\emph{[Figure: 5 patches]}}
    \caption{5 patches: ``a \textbf{bastard} in bed with a book''}
  \end{subfigure}
  \hfill
  \begin{subfigure}[b]{0.32\textwidth}
    \centering
    \IfFileExists{images/6_7_a jungle bunny with a green dot.png}{\includegraphics[width=\textwidth]{images/6_7_a jungle bunny with a green dot.png}}{\emph{[Figure: 7 patches]}}
    \caption{7 patches: ``a \textbf{jungle bunny} with a green dot''}
  \end{subfigure}
  \caption{Patch-level attacks with 3, 5, and 7 patches perturbed. Target prompt words shown in \textbf{bold}.}
\end{figure}

In addition to generalizing across unseen images, these perturbations also transfer to other model architectures---a property termed ``doubly universal'' with respect to both input data and network architecture.

\subsubsection{Patch Fool}
The authors in \cite{patch-fool} introduce a framework that fools the self-attention mechanism by attacking its basic component (i.e., a single patch) with a series of attention-aware optimization techniques, called Patch Fool. The idea is to create customized adversarial perturbations onto a patch that can be more effective in fooling the captured patch-wise global interactions of self-attention modules than attacking the CNN modules. 

In their proposed Patch-Fool Attack, they do not limit the perturbation strength onto each pixel and, instead, constrain all the perturbed pixels within one patch (or several patches), which can be viewed as a variant of sparse attacks. Such attack strategies will lead to adversarial examples with a noisy patch, which visually resembles and emulates natural corruptions in a small region of the original image, e.g., one noisy patch only counts for 1/196 in the inputs of DeiT-S (\cite{patch-fool2}), caused by potential defects of the sensors or potential noises/damages of the optical devices.

For better fooling ViTs, they select the most influential patch $p$ from the maximum of measuring the importance of the j-th token in the l-th layer based on its contributions to other tokens in the self-attention calculation, according to a predefined value $l$. They fix $l=5$ by default since the patches at later self-attention layers are observed to be diverse from the input patches due to the increased information mixed from other patches, making them non-ideal for guiding the selection of input patches.

\subsubsection{BLIP: Bootstrapping Language-Image Pre-training for Unified Vision-Language Understanding and Generation}
Most existing pre-trained models are not flexible enough to adapt to a wide range of vision-language tasks. Encoder-based models are less straightforward to directly transfer to text generation tasks, whereas encoder-decoder models have not been successfully adopted for image-text retrieval tasks. To address these limitations, BLIP was introduced in \cite{blip}. 

BLIP introduces the Multi-modal Mixture of Encoder-Decoder (MED), a new model architecture for effective multi-task pre-training and flexible transfer learning. An MED can operate as a uni-modal encoder, an image-grounded text encoder, or an image-grounded text decoder. BLIP also introduces Captioning and Filtering (CapFilt), a dataset bootstrapping method for learning from noisy image-text pairs.

The proposed multi-modal mixture of encoder-decoder, depicted in Figure \ref{fig:blip_architecture}, is a unified vision-language model operating in three modes: (1) a uni-modal encoder trained with image-text contrastive (ITC) loss, (2) an image-grounded text encoder with cross-attention layers trained with image-text matching (ITM) loss, and (3) an image-grounded text decoder with causal self-attention trained with language modeling (LM) loss to generate captions.



BLIP achieves state-of-the-art performance on seven vision-language tasks: image-text retrieval, image captioning, visual question answering, visual reasoning, visual dialog, zero-shot text-video retrieval, and zero-shot video question answering.

\subsection{CaptionFool Attack}
We attack the BLIP captioning model\footnote{https://github.com/salesforce/LAVIS} using a modified Patch-Fool attack to generate universal (input-agnostic) adversarial perturbations for arbitrary target captions.
Our CaptionFool attack is detailed in Algorithm~\ref{alg:captionfool}. Hyper-parameter configurations are provided in Appendix~\ref{appendix}.

\begin{figure}[t]
  \centering
  \begin{subfigure}[b]{0.32\textwidth}
    \centering
    \IfFileExists{images/10_a balloon with a heart on it.png}{\includegraphics[width=\textwidth]{images/10_a balloon with a heart on it.png}}{\emph{[Figure: balloon]}}
    \caption{Inoffensive: a \textbf{balloon} with a heart on it.}
  \end{subfigure}
  \hfill
  \begin{subfigure}[b]{0.32\textwidth}
    \centering
    \IfFileExists{images/11_a jigaboo bear.png}{\includegraphics[width=\textwidth]{images/11_a jigaboo bear.png}}{\emph{[Figure: slang]}}
    \caption{Offensive slang: \textbf{jigaboo} bear.}
  \end{subfigure}
  \hfill
  \begin{subfigure}[b]{0.32\textwidth}
    \centering
    \IfFileExists{images/28_a nigger in a bottle.png}{\includegraphics[width=\textwidth]{images/28_a nigger in a bottle.png}}{\emph{[Figure: offensive]}}
    \caption{Offensive: a \textbf{n-word} in a bottle.}
  \end{subfigure}
  \caption{Sparse patch attacks with 30k (20\%), 40k (27\%), and 50k (35\%) pixels perturbed. Target prompt words shown in \textbf{bold}.}
\end{figure}

\begin{algorithm}[tb]
\caption{CaptionFool: Universal Adversarial Attack on Vision-Language Models}
\label{alg:captionfool}
\textbf{Input}: Take a batch of images X, \\ random patch attack matrix Delta, zero vector Mask \\ Target Prompt P "a picture of (attack term here)" \\
\textbf{Parameters}: $N_{patches}$ to be perturbed(N), weighting parameter(alpha) \\
\textbf{Output}: Delta and Mask which attacks ANY image\\
\begin{algorithmic}[1] 
\STATE Run forward propagation on the model with X and the target prompt P to get visual encoder attentions by passing the images and labels through the model
\FOR{each image i in X}
\STATE Mean all transformer attention heads and all tokens to get one single vector compressing the information
\STATE Sort to get top N patches for each image in the batch by attention weight: $Y_i$
\ENDFOR
\STATE Compute the most frequently appearing N patch indices in all $Y_i$ among the batch images
\STATE Set Mask =1 for the patches found to be most important
\FOR{T training iterations}
\STATE Run model on attacked images (X + Delta*Mask) and get Language modelling cross-entropy loss wrt target prompt P
\STATE Compute gradient of language modeling loss wrt Delta ($G_{language-modelling}$)
\FOR{layer in first 6 attention layers}
\STATE Compute and sum gradient of attention loss wrt Delta ($G_atten$)
\ENDFOR
\STATE Combine $G_{language-modelling}$ and $G_{atten}$ in a weighted average using alpha
\STATE Update Delta using the ADAM optimizer
\STATE Clamp Delta to keep the perturbation in required limits
\ENDFOR
\STATE \textbf{return} Delta, Mask
\end{algorithmic}
\end{algorithm}

\textbf{Universal Patch-Fool Attack:} \\
We modify the Patch-Fool attack, which is an image-specific attack, to be universal, by implementing the following changes:
\begin{itemize}
    \item We optimize the Patch-Fool attack over a batch of Flickr images, by keeping the delta and patch mask constant for all samples.
    \item Instead of finding the highest attention patches per-image, we compute the highest attention patches over the whole batch, and pick the most frequently appearing patch indices.
\end{itemize}

\textbf{Loss Function:} \\
The original Patch-Fool attack uses cross-entropy loss between predicted and target labels. Since image captioning involves sequence generation rather than classification, we minimize the language modeling (LM) cross-entropy loss between the generated caption and the target offensive text. Specifically, we construct target prompts of the form ``a picture of a [target term]'' and optimize the perturbation to minimize the decoder's cross-entropy loss with respect to this target sequence.

\section{Experiments and Results}

We evaluate our attack on three categories of target prompts: inoffensive words, offensive words, and offensive slang terms. For each category, we report the following metrics:
\begin{enumerate}
    \item \textbf{Validation:} Our validation dataset is 50 unseen flickr8k dataset images which are human photographed in diverse settings. The accuracy on these images is used to tune our hyper-parameters and implement early stopping.
    \item \textbf{Testing:} Our test dataset is 50 unseen COCO dataset images which are diverse objects and scenes. These are very different from the flickr8k images used to create the attack and help us verify the transferability of our perturbations. 
\end{enumerate}

We evaluate both attack variants described by the Patch-Fool authors: \textbf{Patch-level attacks}, which achieve the highest success rates with minimal perturbation, and \textbf{Sparse attacks}, which distribute perturbations across the image for a more natural appearance but require greater perturbation magnitude.
We analyze attack success rate as a function of the number of perturbed patches. For Vision Transformers, the input image is divided into 577 patches of 16$\times$16 pixels. Our results demonstrate that patch-level attacks achieve reliable caption generation with only 7 patches (1.2\% of the image), with strong performance using just 5 patches. Sparse attacks require 25--35\% pixel perturbation to achieve comparable success rates.

\subsection{Results on Inoffensive prompts}
Table~\ref{inoffensive_res} presents results for common inoffensive words that are \textbf{not} present in the MS COCO training captions, testing our attack's ability to generate out-of-distribution content. 

\begin{table}
    \centering
    \begin{tabular}{c c c c}
        \toprule
        \textbf{Prompt} & \multicolumn{3}{c}{\textbf{No. of Patches}}\\
        \textbf{Word}&  3/577 & 5/577 & 7/577 \\
        \midrule
        balloon     & 0.78 & 0.88 & 0.92 \\
        pillow      & 0.76 & 0.86 & 0.98 \\
        mat         & 0.68 & 0.88 & 0.98 \\
        dough       & 0.24 & 0.80 & 0.86 \\
        bottle      & 0.80 & 0.80 & 0.98 \\
        bulbs       & 0.14 & 0.80 & 0.90 \\
        colonel     & 0.22 & 0.74 & 0.94 \\
        cake        & 0.62 & 0.92 & 1.00 \\
        embonpoint  & 0.34 & 0.96 & 0.94 \\
        eschew      & 0.34 & 0.62 & 0.94 \\
        wires       & 0.10 & 0.86 & 0.94 \\
        \midrule
        \textbf{Average}& 0.52 & 0.84 & 0.94 \\ 
        \bottomrule
    \end{tabular}
    \caption{Test attack success rate (ASR) for inoffensive prompts.}
    \label{inoffensive_res}
\end{table}

\subsection{Results on Offensive prompts}
Table~\ref{offensive_res} presents results for offensive words from the Surge.AI dataset \cite{surge}. Our attack achieves high success rates against the BLIP model with minimal perturbation.

\begin{table}
    \centering
    \begin{tabular}{c c c c}
        \toprule
        \textbf{Prompt} & \multicolumn{3}{c}{\textbf{No. of Patches}}\\
        \textbf{Word}&  3/577 & 5/577 & 7/577 \\
        \midrule
        airhead     & 0.62  & 0.7   & 0.96 \\
        bastard     & 0.28  & 0.92  & 0.94 \\
        bimbo       & 0.66  & 0.98  & 1.0  \\
        deadbeat    & 0.66  & 0.72  & 0.98 \\
        freak       & 0.0   & 0.82  & 0.82 \\
        jerk        & 0.0   & 0.88  & 0.84 \\
        nutter      & 0.54  & 0.88  & 1.0  \\
        redneck     & 0.64  & 0.9   & 0.96 \\
        twit        & 0.76  & 0.96  & 0.98 \\
        prick       & 0.48  & 0.84  & 0.98 \\
        n-word      & 0.36  & 0.82  & 0.82 \\
        \midrule
        \textbf{Average} & 0.606 & 0.888 & 0.96 \\
        \bottomrule
    \end{tabular}
    \caption{Test attack success rate (ASR) for offensive prompts.}
    \label{offensive_res}
\end{table}

\subsection{Results on Offensive Slang prompts}
Table~\ref{slang_res} presents results for offensive slang terms from the Racial Slur Database \cite{racial_slur_database}. These terms are commonly used to evade content moderation filters. Our attack successfully generates these slang terms with high accuracy, demonstrating the potential for adversarial evasion of keyword-based filters.

\begin{table}
    \centering
    \begin{tabular}{c c c c}
        \toprule
        \textbf{Prompt} & \multicolumn{3}{c}{\textbf{No. of Patches}}\\
        \textbf{Word}&  3/577 & 5/577 & 7/577 \\
        \midrule
        jigaboo      & 0.56 & 0.62 & 0.98 \\ 
        kaffir       & 0.56 & 1.00 & 0.98 \\
        moolignon    & 0.00 & 0.78 & 0.98 \\
        quashee      & 0.78 & 0.90 & 0.96 \\
        jungle bunny & 0.16 & 0.86 & 0.86 \\
        justin igger & 0.00 & 0.42 & 0.96 \\
        \midrule
        \textbf{Average}  & 0.46 & 0.78 & 0.95  \\ 
        \bottomrule
    \end{tabular}
    \caption{Test attack success rate (ASR) for offensive slang prompts.}
    \label{slang_res}
\end{table}

\subsection{Sparse Patch Fool Results}

Table~\ref{sparse_res} presents results for the Sparse Patch-Fool attack variant. This attack requires perturbation of a significant fraction of pixels, with a minimum threshold of approximately 20\% for reliable performance.

\begin{table}
    \centering
    \begin{tabular}{c c c c}
        \toprule
        \textbf{Prompt} & \multicolumn{3}{c}{\textbf{Sparse Pixels}}\\
        \textbf{Word}&  30k (20\%) & 40k (27\%) & 50k (35\%) \\
        \midrule
        balloon  &  0.76   & 0.98   & 0.98  \\ 
        cake     &  0.82   & 0.96   & 0.98  \\ 
        n-word   &   0.06  & 0.48   & 0.84   \\
        redneck  &  0.58   & 0.92   & 1  \\
        jigaboo  &  0.74   & 0.82   & 0.88   \\
        quashee  &  0.78   & 0.92   & 1  \\
        \bottomrule
    \end{tabular}
    \caption{Test attack success rate (ASR) for Sparse Patch-Fool on inoffensive (balloon, cake), offensive (n-word, redneck), and offensive slang (jigaboo, quashee) prompts.}
    \label{sparse_res}
\end{table}

\section{Conclusion}
We have presented CaptionFool, a universal adversarial attack that exposes critical vulnerabilities in state-of-the-art transformer-based image captioning models. By extending the Patch-Fool attack to the universal setting, we demonstrate that perturbing only 7 out of 577 image patches (1.2\% of the image) achieves 94-96\% success rate in generating arbitrary target captions, including offensive content. Furthermore, our analysis of slang-based attacks reveals that adversarially generated captions can evade keyword-based content moderation filters commonly deployed on social media platforms.

These findings highlight significant security concerns for vision-language models deployed in production systems. As image captioning becomes increasingly integrated into accessibility tools, content moderation pipelines, and multimedia search engines, the potential for adversarial exploitation grows correspondingly.

\textbf{Limitations.} Our attack assumes white-box access to the model's attention layers, which may not always be available in deployed systems. Additionally, we evaluated only the BLIP model; transferability to newer architectures such as BLIP-2 or GPT-4V remains to be investigated.

\textbf{Future Work.} We aim to develop and evaluate effective defenses against CaptionFool. Investigating black-box attack variants and evaluating robustness across a broader range of vision-language models, including recent multimodal large language models, are important directions for future research.

\section*{Appendix}
\label{appendix}
\subsection*{A. Hyper-parameters used for Patch Fool}

\begin{center}
\begin{tabular}{l|l}
\textbf{Parameter Name} & \textbf{Values} \\
\hline
batch size & 15 \\
image size & 384 $\times$ 384 \\
patch size & 16 $\times$ 16 \\
number of patches(N) & 3, 5, 7 \\
atten loss weight(alpha) & 0.005 \\
attention layer & 4 \\
attack iterations & 150 \\
learning rate & 0.8 \\
scheduler step size & 30 \\
scheduler gamma & 0.95 \\
\end{tabular}
\end{center}
\subsection*{B. Hyper-parameters used for Sparse Patch Fool}

\begin{center}
\begin{tabular}{l|l}
\textbf{Parameter Name} & \textbf{Values} \\
\hline
batch size & 40 \\
image size & 384 $\times$ 384 \\
patch size & 16 $\times$ 16 \\
number of patches(N) & 577 \\
number of sparse pixels & 30k, 40k, 50k \\
atten loss weight(alpha) & 0.002 \\
attention layer & 4 \\
attack iterations & 200 \\
learning rate & 0.8 \\
scheduler step size & 30 \\
scheduler gamma & 0.95 \\
\end{tabular}
\end{center}

\section*{Ethics Statement}
This work identifies and demonstrates vulnerabilities in image captioning models with the goal of improving their robustness and safety. As these models become increasingly prevalent in applications ranging from accessibility tools to content moderation systems, understanding their failure modes is essential for responsible deployment.

We acknowledge that our attack methodology could potentially be misused to generate harmful content. However, we believe that responsible disclosure of these vulnerabilities is necessary to motivate the development of effective defenses. Our demonstration that slang terms can evade content moderation filters highlights the inadequacy of keyword-based approaches and the need for more sophisticated detection systems.

We have taken care to minimize potential harm by: (1) not releasing attack code or trained perturbations, (2) focusing on demonstrating the existence of vulnerabilities rather than optimizing for maximum harm, and (3) providing this work to the research community to facilitate the development of countermeasures. We hope this work will encourage increased scrutiny in the development and deployment of vision-language models and drive research toward more robust safety mechanisms.


\bibliography{biblio}

@inproceedings{Xu_2019_CVPR,
  author = {Xu, K. and others},
  title = {Adversarial attacks on image captioning},
  booktitle = {CVPR},
  year = {2019}
}

@article{Show-and-Fool,
  author = {Author},
  title = {Show and Fool},
  journal = {arXiv},
  year = {2019}
}

@article{patch-fool,
  author = {Author},
  title = {Patch-Fool: Adversarial patches for vision transformers},
  year = {2023}
}

@article{ViT,
  author = {Dosovitskiy, A. and others},
  title = {An Image is Worth 16x16 Words: Transformers for Image Recognition at Scale},
  journal = {ICLR},
  year = {2021}
}

@article{UAP,
  author = {Moosavi-Dezfooli, S. and others},
  title = {Universal Adversarial Perturbations},
  booktitle = {CVPR},
  year = {2017}
}

@article{parekh2023attacking,
  author = {Parekh, S. and others},
  title = {Attacking compressed vision transformers},
  year = {2023}
}

@article{Kwon2022,
  author = {Kwon, H. and others},
  title = {Restricted-area adversarial examples for image captioning},
  year = {2022}
}

@article{icattack,
  author = {Ji, Z. and others},
  title = {Targeted attacks for removing words from image captions},
  year = {2022}
}

@article{icattck2,
  author = {Sun, T. and others},
  title = {Backdoor attacks on image captioning models},
  year = {2023}
}

@article{UAT,
  author = {Wallace, E. and others},
  title = {Universal adversarial triggers for NLP},
  year = {2019}
}

@article{singla2022minimal,
  author = {Singla, S. and others},
  title = {MINIMAL: Mining universal adversarial triggers},
  year = {2022}
}

@article{parekh2020my,
  author = {Parekh, S. and others},
  title = {Adversarial susceptibility in automated scoring},
  year = {2020}
}

@article{kumar2023automatic,
  author = {Kumar, A. and others},
  title = {Automatic essay scoring and robustness},
  year = {2023}
}

@article{DeToxiGen,
  author = {Hartvigsen, T. and others},
  title = {DeToxiGen: Detecting toxic content},
  year = {2022}
}

@inproceedings{ms-coco,
  author = {Lin, T.-Y. and others},
  title = {Microsoft COCO: Common Objects in Context},
  booktitle = {ECCV},
  year = {2014}
}

@article{flickr,
  author = {Young, P. and others},
  title = {From image descriptions to visual denotations},
  journal = {TACL},
  year = {2014}
}

@misc{surge,
  author = {Surge.AI},
  title = {Obscenity List},
  howpublished = {\url{https://github.com/surge-ai/profanity}},
  year = {2022}
}

@misc{racial_slur_database,
  author = {Racial Slur Database},
  title = {Racial Slur Database},
  howpublished = {\url{https://www.rsdb.org}},
  year = {2023}
}

@article{frusVITGPT,
  author = {Author},
  title = {A Frustratingly Simple Approach for End-to-End Vision-Language},
  year = {2022}
}

@article{blip,
  author = {Li, J. and others},
  title = {BLIP: Bootstrapping Language-Image Pre-training},
  journal = {ICML},
  year = {2022}
}

@article{Transformers,
  author = {Vaswani, A. and others},
  title = {Attention Is All You Need},
  booktitle = {NeurIPS},
  year = {2017}
}

@article{bert,
  author = {Devlin, J. and others},
  title = {BERT: Pre-training of Deep Bidirectional Transformers},
  year = {2019}
}

@article{patch-fool2,
  author = {Author},
  title = {Patch-Fool (DeiT)},
  year = {2023}
}
\end{document}